\documentclass{IEEEtaes}

\usepackage{color,array,amsthm}
\usepackage{graphicx}

\usepackage{amssymb} 
\usepackage{amsmath} 

\usepackage{algorithm} 
\usepackage{algpseudocode} 

\usepackage{color,soul} 

\usepackage{makecell} 

\usepackage{hyperref}
\usepackage{longtable}

\usepackage{xcolor}
\usepackage{multicol}
\definecolor{DGreen}{RGB}{0, 0, 0}  

\jvol{XX}
\jnum{XX}
\jmonth{XXXXX}
\paper{1234567}
\pubyear{2020}
\doiinfo{TAES.2020.Doi Number}

\newcommand{\rom}[1]{%
  \textup{\uppercase\expandafter{\romannumeral#1}}%
}
\newcommand{\eunit}{\text{W} \cdot \text{m}^{-2} \cdot \mu\text{m}^{-1}}
\newcommand{\eunitPh}{\text{photons} \cdot \text{s}^{-1} \cdot\text{m}^{-2} \cdot \mu\text{m}^{-1} }

\begin{document}

\title{Real-Time Convolutional Neural Network-Based Star Detection and Centroiding Method for CubeSat Star Tracker}

\author{HONGRUI ZHAO}
\author{MICHAEL F. LEMBECK}
\author{ADRIAN ZHUANG}
\author{RIYA SHAH}
\author{JESSE WEI}
\affil{University of Illinois Urbana-Champaign, IL 61820, USA}

\receiveddate{This article has been accepted for publication in IEEE Transactions on Aerospace and Electronic Systems. This is the author's version which has not been fully edited and content may change prior to final publication. Citation information: DOI 10.1109/TAES.2025.3542744}


\authoraddress{Hongrui Zhao, Michael F. Lembeck, Adrian Zhuang, Riya Shah, Jesse Wei are with the Aerospace Engineering Department, University of Illionis Urbana-Champaign, IL 61820, USA
(e-mail: \href{mailto:hongrui5@illinois.edu}{hongrui5@illinois.edu}; \href{mailto:mlembeck@illinois.edu}{mlembeck@illinois.edu}; \href{mailto:adrianz3@illinois.edu}{adrianz3@illinois.edu}; \href{mailto:riyahs3@illinois.edu}{riyahs3@illinois.edu}; \href{mailto:jwei28@illinois.edu}{jwei28@illinois.edu}). \itshape(Corresponding author: Hongrui Zhao.)}

\supplementary{© 20xx IEEE. Personal use of this material is permitted. Permission from IEEE must be obtained for all other uses, in any current or future media, including reprinting/republishing this material for advertising or promotional purposes, creating new collective works, for resale or redistribution to servers or lists, or reuse of any copyrighted component of this work in other works.}

\markboth{H. ZHAO ET AL.}{\MakeUppercase{Real-Time CNN-Based Star Detection and Centroiding Method for CubeSat Star Tracker}}
{An Author: A Title} 
\maketitle

\begin{abstract}
Star trackers are one of the most accurate celestial sensors used for absolute attitude determination. The devices detect stars in captured images and accurately compute their projected centroids on an imaging focal plane with subpixel precision. Traditional algorithms for star detection and centroiding often rely on threshold adjustments for star pixel detection and pixel brightness weighting for centroid computation. However, challenges like high sensor noise and stray light can compromise algorithm performance. This article introduces a Convolutional Neural Network (CNN)-based approach for star detection and centroiding, tailored to address the issues posed by noisy star tracker images in the presence of stray light and other artifacts.
Trained using simulated star images overlayed with real sensor noise and stray light, the CNN produces both a binary segmentation map distinguishing star pixels from the background and a distance map indicating each pixel's proximity to the nearest star centroid. 
Leveraging this distance information alongside pixel coordinates transforms centroid calculations into a set of trilateration problems solvable via the least squares method.
Our method employs efficient UNet variants for the underlying CNN architectures, and the variants' performances are evaluated.
Comprehensive testing has been undertaken with synthetic image evaluations, hardware-in-the-loop assessments, and night sky tests. The tests consistently demonstrated that our method outperforms several existing algorithms in centroiding accuracy and exhibits superior resilience to high sensor noise and stray light interference. An additional benefit of our algorithms is that they can be executed in real-time on low-power edge AI processors.
\end{abstract}

\begin{IEEEkeywords}
Convolution, 
Satellite navigation systems,
Training 
\end{IEEEkeywords}

\section{INTRODUCTION}
With the availability of compact star tracker technology, space missions demanding precise attitude determination are increasingly leveraging these devices on cost-effective CubeSat platforms. 
Laser communication CubeSats in Low Earth Orbit (LEO), when equipped with a star tracker, can achieve precise pointing capabilities towards ground stations, facilitating downlink rates of up to 200 Gbps \cite{bib1}. For deep space CubeSat missions, star trackers can also play a crucial role in high-gain antenna pointing \cite{bib2}.  Star trackers can be used in conjunction with other sensors to accurately determine the orbits of other orbiting objects, enhancing space situational awareness \cite{bib3}.

Many CubeSat star trackers use commercial off-the-shelf (COTS) components, trading cost for performance.  Star trackers equipped with low-cost imaging sensors \cite{cubestar} suffer from lower sensitivity resulting in noisier star images.  Conventional star centroiding algorithms \cite{Liebe}\cite{Gaussian} struggle with high sensor noise, leading to compromised centroid computation accuracy that is pivotal for optimal star tracker performance. Noise can also introduce false positive star detections affecting attitude determination. 

Stray light from bright bodies or optical path dust reflections can also result in false positives. 
Given that the intensity of stray light often surpasses the background brightness level, conventional threshold-based star detection algorithms \cite{Liebe}\cite{WITM}\cite{ST16}\cite{erosion} are unable to discriminate false detections, undermining subsequent star identification processes. 
To counteract stray light, large baffles are frequently placed in front of the sensor optics \cite{ST16Data}, potentially encroaching on the limited payload space available in CubeSats. 
Even with a baffle, star trackers remain susceptible to blinding when directly exposed to bright bodies in their field of view \cite{MarCO}.

In this article, we tackle these deficiencies using Convolutional Neural Networks (CNNs). 
CNNs have demonstrated remarkable image processing capabilities relevant to star trackers across a diverse set of applications, including precision agriculture, where CNNs excel in precisely locating and counting crops from aerial images \cite{centroidnet}. 
For low-light imaging scenarios, CNNs effectively denoise and enhance short-exposure images characterized by low photon counts and diminished signal-to-noise (SNR) \cite{dark}. 
Recent advancements in star tracker research have leveraged CNNs for star image segmentation, showcasing robust performance against high sensor noise and stray light \cite{italy}\cite{SAA-UNet}. 
Image segmentation facilitates star pixel detection, but a subsequent denoising step is still required to achieve high SNR star images for accurate centroiding. 
Adopting two distinct CNNs, one for star detection and another for denoising, imposes significant computational overhead, making it impractical for edge computing hardware deployment. 
To overcome these limitations, this article advocates for the end-to-end training of a singular CNN capable of simultaneously handling star detection and centroiding, resulting in a streamlined and computationally efficient architecture.

The main contributions of this article are:
\begin{enumerate}
  \def\labelenumi{\arabic{enumi})}
  \item
    We introduce an end-to-end training approach for a singular CNN, designed to concurrently handle star detection and centroiding.
    \textcolor{DGreen}{This represents a significant advancement over existing methods, which typically focus on either detection or centroiding in isolation.}
  \item
      \textcolor{DGreen}{We introduce a method for generating synthetic star images with realistic stray light and sensor noise. This approach minimizes the reliance on labor-intensive manual labeling of real star images, making our pipeline more efficient and scalable. The introduction of stray light in training data allows our method to achieve superior detection robustness.}
  \item
    \textcolor{DGreen}{We propose a new centroid estimation technique that leverages a distance map and trilateration, offering improved accuracy compared to conventional intensity-based methods.}
  \item 
    We conduct a comprehensive performance evaluation of various CNN models and identifying the most suitable CNN architectures for real-time star tracker image processing.
  \item 
    \textcolor{DGreen}{We conduct a series of rigorous experiments with synthetic data, hardware simulator, and night sky test to evaluate the performance of our method. }
\end{enumerate}

This article is structured as follows:
In Section \rom{2}, we introduce traditional star detection and centroiding algorithms, as well as existing CNN models relevant to this work.
Section \rom{3} presents the workflow of the proposed star detection and centroiding method, the data generation process, and the CNN implementation details. 
Section \rom{4} explains the experiment setups and discusses the experiment results from synthetic dataset testing, hardware-in-the-loop testing, and night sky testing.
Finally, we conclude this article and point out the future works in Section \rom{5}.

\section{Related Works}
Existing star detection algorithms can be broadly classified into two categories: global threshold methods, and local threshold methods. 
Global threshold methods determine a single pixel intensity threshold to distinguish star pixels (foreground) from the background.
Liebe's global adaptive threshold method \cite{Liebe} computes the mean and the standard deviation of each star image, where the standard deviation is scaled and added to the mean to derive the global threshold.
While Liebe's method offers a straightforward and effective detection approach, manual tuning of the scaling factor is often required to achieve optimal performance.
Another approach, using the weighted iterative threshold method (WITM) \cite{WITM}, computes a global threshold for each star image through iterative adjustments of two weight coefficients until the threshold difference between successive iterations falls below a predefined value.
Although WITM can autonomously derive an optimal threshold and exhibits reduced sensitivity to initialization, its iterative nature considerably impacts computation speed.
On the other hand, local threshold methods assign individual thresholds to each pixel, necessitating each pixel to surpass its designated threshold to be classified as a star pixel.
The detection routine of the ST-16 star tracker \cite{ST16} employs a $1 \times 29$ window centered on each pixel and utilizes the window's average, augmented by a predetermined constant, as the local threshold.
Meanwhile, the method proposed by Sun \emph{et al.} \cite{erosion} employs a sequential application of a Gaussian filter, followed by erosion, dilation, and an average filter on each star image to compute a background estimate.
For a pixel to qualify as a star pixel, its intensity must exceed the computed background estimate plus a constant.
Although this method demonstrates robustness against uneven background illumination, it requires considerable computations due to the sequential application of multiple filters.

Threshold-based detection methods often fail to distinguish faint stars from background noise or discriminate between stars and other bright non-stellar objects.
To enhance the robustness of star detection, several approaches have leveraged heuristic criteria \cite{ST16} and local gradient information \cite{moon}.
In the ST-16 detection routine \cite{ST16},  a cluster of pixels surpassing their respective local thresholds is identified. The algorithm then quantifies the number of connected bright pixels within the cluster using 4-connectivity. Subsequently, both the count of contiguous bright pixels and their cumulative intensity are used to eliminate isolated hot pixels.
Jiang \emph{et al.} \cite{moon} suggested that gradient information from lunar images could be examined and used to manually tune gradient thresholds to differentiate stars from lunar interference. 
Despite their merits, these aforementioned techniques exhibit sensitivity to variations in environmental conditions and sensor characteristics, necessitating frequent parameter recalibrations.
In contrast, our method addresses this limitation by incorporating training images spanning a range of exposure durations and noise levels.
This design allows the neural network to adapt dynamically to varying imaging conditions without the need for recurrent recalibrations.

\begin{figure*}
  \centerline{\includegraphics[trim={1.5cm 2cm 6cm 1.5cm},clip,width=38pc]{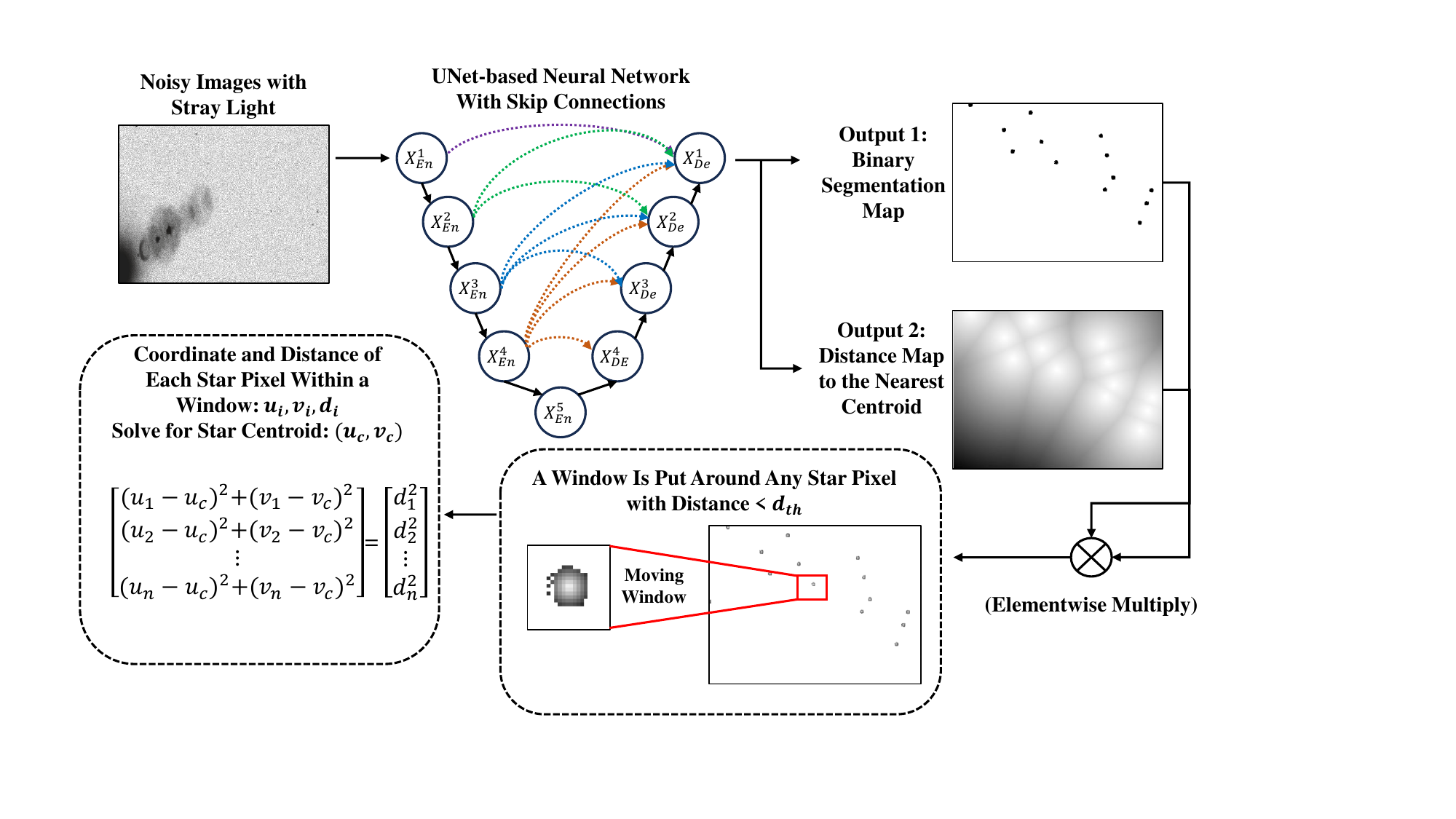}} 
  \caption{Flowchart of our star detection and centroiding method. 
  Noisy star images with stray light are fed into a UNet-based CNN. 
  The neural network outputs both a binary segmentation map of star pixels versus background and a distance map of each pixel to the nearest star centroid.
  Using the distance information and the pixel coordinates, star centroiding now becomes a trilateration problem and can be directly solved using a least squares solution.}
  \label{fig1:flowchart}
\end{figure*}

Existing star-centroiding algorithms utilize the intensity distribution of star pixels to compute star centroids with subpixel accuracy. 
The center of gravity method stands out as a common approach due to its simplicity and computational efficiency \cite{Liebe}. 
This method derives star centroids by averaging the coordinates of a cluster of star pixels, each weighted by its brightness. 
However, the center of gravity method yields only an approximate centroid position and proves susceptible to background noise.
In pursuit of enhanced accuracy and robustness, Delabie \emph{et al.} introduced the Gaussian grid algorithm \cite{Gaussian}. 
This approach employs a Gaussian distribution function to model the detected star pixels within a designated window, subsequently deriving the centroid as the center of this distribution. 
The Gaussian grid algorithm surpasses the center of gravity method in terms of accuracy and robustness while maintaining computational efficiency.
However, the Gaussian grid algorithm can fail when confronted with strongly non-Gaussian distributions of star pixels, such as in the presence of stray light or sensor motion.
In contrast, our method is trained for both star detection and centroiding. This integrated approach ensures robust centroiding performance across varying noise levels, thereby enhancing overall accuracy and robustness.

The UNet, originally designed for medical image segmentation, incorporates skip connections to retain fine details from the input image \cite{UNet}.
These skip connections merge feature maps containing high spatial frequency details from early convolutional layers with those containing high-level semantic information. 
While UNet has demonstrated impressive performance in star image segmentation with a notably low false positive rate \cite{italy}\cite{SAA-UNet}, its original architecture, comprising approximately 31 million parameters, demands significant computational resources and executes in seconds on a standard personal computer. 
This computational intensity poses a challenge for real-time star tracker processing on a low-power processor. 
To address this limitation, various efforts have been undertaken to enhance UNet's computational efficiency and reduce its model size. 
MobileUNet \cite{MobileUNet} integrates inverted residual blocks from MobileNetV2 to replace the VGG-16 blocks in conventional U-Net, resulting in marked computational efficiency and a reduction in model size. 
ELUNet \cite{ELUNet} introduces deep skip connections that effectively reduces channel numbers without sacrificing performance. 
Finally, SqueezeUNet \cite{SqueezeUNet} employs fire modules to decrease the number of channels, offsetting this reduction with an inception stage comprising two parallel convolutions. 
In this article, we undertake an exhaustive analysis of these UNet variants to determine the most suitable architectures for star tracker applications.

\section{PROPOSED METHOD}
Fig. \ref{fig1:flowchart} illustrates the flowchart detailing our CNN-based method for star detection and centroiding. 
The neural network receives input star images that may contain stray light and varying noise levels. 
Leveraging a UNet-based CNN, our method produces both a binary segmentation map and a distance map. 
Within the segmentation map, a ``one'' indicates a star pixel has been detected, while a ``zero'' signifies a background pixel.
Concurrently, the distance map assigns each pixel a value corresponding to its distance from the nearest star centroid. 
These two maps are combined using element-wise multiplication. 
In the resultant map, pixels with values below a specified threshold, $d_{th}$, suggest potential star centroid locations, prompting the centering of a window at each corresponding pixel.
Within each window, every pixel $i$ is identified by its coordinates $(u_i, v_i)$ and stores its distance $d_i$ to the star centroid $(u_c, v_c)$.
This transforms star centroiding into a trilateration problem, determining a position based on distances from at least three known points, that can be resolved using a least squares solution. 
The following sections will delve into the generation of training data, the procedure for star detection and centroiding, and the specifics of our CNN implementation.

\subsection{Training Data Generation}
To generate training images for the neural network, the initial step involves acquiring synthetic noiseless star images.
The Hipparcos catalog \cite{hipparcos} is used to provide the position and magnitude information of stars.
Considering the camera parameters and the actual orientation of a star tracker, determining actual centroids and identifications of stars within the image plane is straightforward. 
Subsequent efforts accurately capture the distribution patterns and magnitudes of real stars.

The Hipparcos catalog provides the Johnson V magnitude in the Johnson UBV photometric system.
The V band filter is centered at $\lambda_0 = 0.55 \,\mu\text{m}$ with a full width at half maximum (FWHM) bandwidth of $\Delta \lambda_0 = 0.089\,\mu\text{m} $, and has the zero magnitude spectral flux density $e_0 = 3.92 \times 10^{-8} \, \eunit$ \cite{astro}.
Therefore, we can obtain the average spectral flux density of a star over the V band in $\eunitPh$ when its magnitude $m_V$ from the Hipparcos catalog is given 
\begin{equation}
  \bar{e}_V = \frac{e_0 \lambda_0}{h \cdot c} \cdot 10^{-\left(\frac{m_V}{2.5}\right)} = 1.085356 \times 10^{11} \cdot 10^{-\left(\frac{m_V}{2.5}\right)}
\end{equation}
where $h$ is the Planck constant and $c$ is the speed of light.
For a star with a magnitude $m_V$, the average photon arrival rate received by an image sensor over the V band is \cite{photonCal}
\begin{equation}
    \label{photon}
    \bar{n} = \bar{e}_V \cdot AP \cdot \bar{\epsilon} \cdot \bar{E} \cdot \Delta \lambda_0
\end{equation}
where $AP$ is the area of the image sensor aperture given in $\text{m}^2$, $\bar{\epsilon}$ is the average optical efficiency factor over V band,
$\bar{E}$ is an average correction factor for extinction due to the airmass over the V band. For on-orbit star tracker processing, $\bar{E}=1$.
Also recall that $\Delta \lambda_0 = 0.089\,\mu\text{m} $ for the V band.

\begin{figure}
  \centerline{\includegraphics[trim={10cm 5cm 12cm 2cm},clip,width=18.5pc]{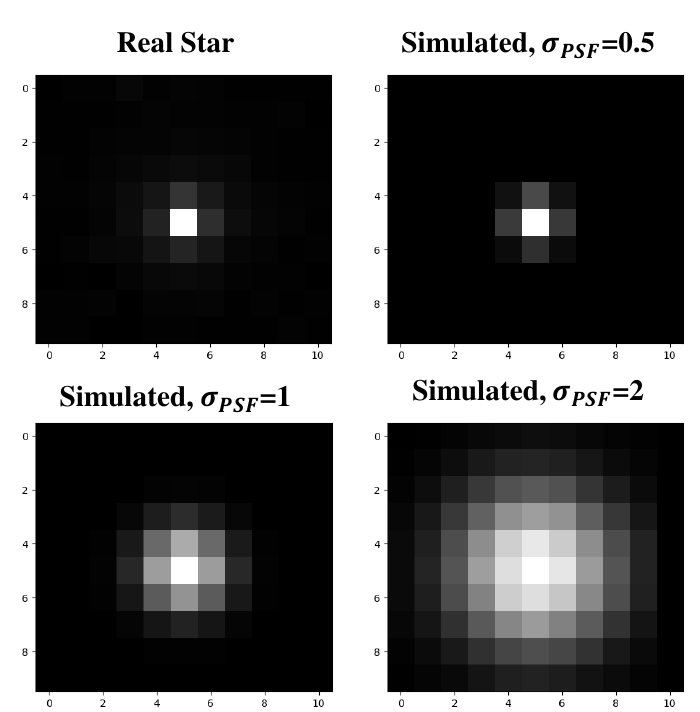}} 
  \caption{Our experiments find that uniformly sampling $\sigma_{\text{psf}}$ from $[0.5,1]$ produces simulated star images with realistic distribution patterns.}
  \label{fig2:sigma}
\end{figure}

The image of a star is subject to optical aberrations and spreads over several pixels, which is modeled by a point spread function (PSF).
The Gaussian distribution serves as a favorable approximation to the PSF of an ideal lens. 
It is expressed in both the width and height directions across the image plane, as described in \cite{Gaussian}
\begin{align}
    g_{width}(u) &= \dfrac{1}{\sqrt{2\pi}\sigma_{\text{psf},u}}\exp\left[ -\dfrac{(u-u_c)^2}{2\sigma_{\text{psf},u}^2 } \right]\\
    g_{height}(v) &= \dfrac{1}{\sqrt{2\pi}\sigma_{\text{psf},v}}\exp\left[ -\dfrac{(v-v_c)^2}{2\sigma_{\text{psf},v}^2 } \right]
\end{align}
where the coordinates of the real star centroid are $u_c$, $v_c$. $\sigma_{\text{psf},u}$ and $\sigma_{\text{psf},v}$ are the PSF width in width and height direction in pixels. We choose $\sigma_{\text{psf},u} = \sigma_{\text{psf},v} = \sigma_{\text{psf}}$ in our simulation.
\textcolor{DGreen}{Following \cite{Gaussian}\cite{PSF}, we assume an ideal aberration-free lens with symmetric defoucsing pattern.} 
In our experiments, we established that uniformly sampling $\sigma_{\text{psf}}$ from $[0.5,1]$ produces star images with realistic distribution patterns, as illustrated in Fig. \ref{fig2:sigma}.
The PSF of a real star captured by a camera set to focus at infinity typically falls within this range.
We also found that training the neural network with a variety of $\sigma_{\text{psf}}$ values can improve its capacity to process stars distorted by \textcolor{DGreen}{lens aberrations (such as coma)} or motion, which may \textcolor{DGreen}{produce asymmetric star images}.

Using the Gaussian PSF, the average rate of arrival of photons received by a pixel with coordinate $(u_i,v_i)$ can be computed as
\begin{align}
    \bar{n}_{pixel} &=  \bar{n} \int_{u_i-0.5}^{u_i+0.5} g_u(u) \,du \int_{v_i-0.5}^{v_i+0.5} g_v(v) \,dv \nonumber 
\end{align}
where each pixel has a non-dimensional width of 1. 

The arrival of photons is a stochastic process $\boldsymbol{x} (t)$ made up of Poisson impulses \cite{astro}. 
The probability of a pixel obtaining $p$ photons during a exposure time $T$ (In our data generator, we uniformly sample $T$ from $[0.1s, 1s]$) is given by 
\begin{equation}
    P\left[ \int_{T} \boldsymbol{x}(t) \,dt = p \right] = \dfrac{ (\bar{n}_{pixel}T)^{p} \, e^{-\bar{n}_{pixel}T} }{p!}
\end{equation}
Sampling from this Poisson distribution will give us the total number of photons $N$ received by the pixel $(u_i,v_i)$ within the exposure time.

Finally, the pixel bit counts $V_{(u_i,v_i)}$ can be calculated \cite{Gaussian}
\begin{equation}
  V_{(u_i,v_i)} = floor\left( N \cdot QE \cdot \frac{2^{N_{bits}}-1}{FWC} \right) \cdot \frac{FWC}{2^{N_{bits}}-1}
\end{equation}
where $floor()$ outputs the greatest integer less than or equal to the input value, $QE$ is the quantum efficiency of the image sensor, $FWC$ is the full well capacity of the image sensor in electrons, and $N_{bits}$ is the
resolution of the ADC. 
A window with $\text{half width} = floor \left( 3\sigma_{\text{psf}} + 0.5 \right)$ centered at the centroid pixel is used to simulate the defocused star image. 
Light contribution $V_{(u_i,v_i)}$ is calculated for every pixel within the window.
\textcolor{DGreen}{We use this window to identify pixels belonging to a star and generate binary segmentation maps. Pixels outside this window are excluded due to their distance from the star's center.}

\begin{figure}
  \centerline{\includegraphics[trim={2cm 3.5cm 8cm 2cm},clip,width=20pc]{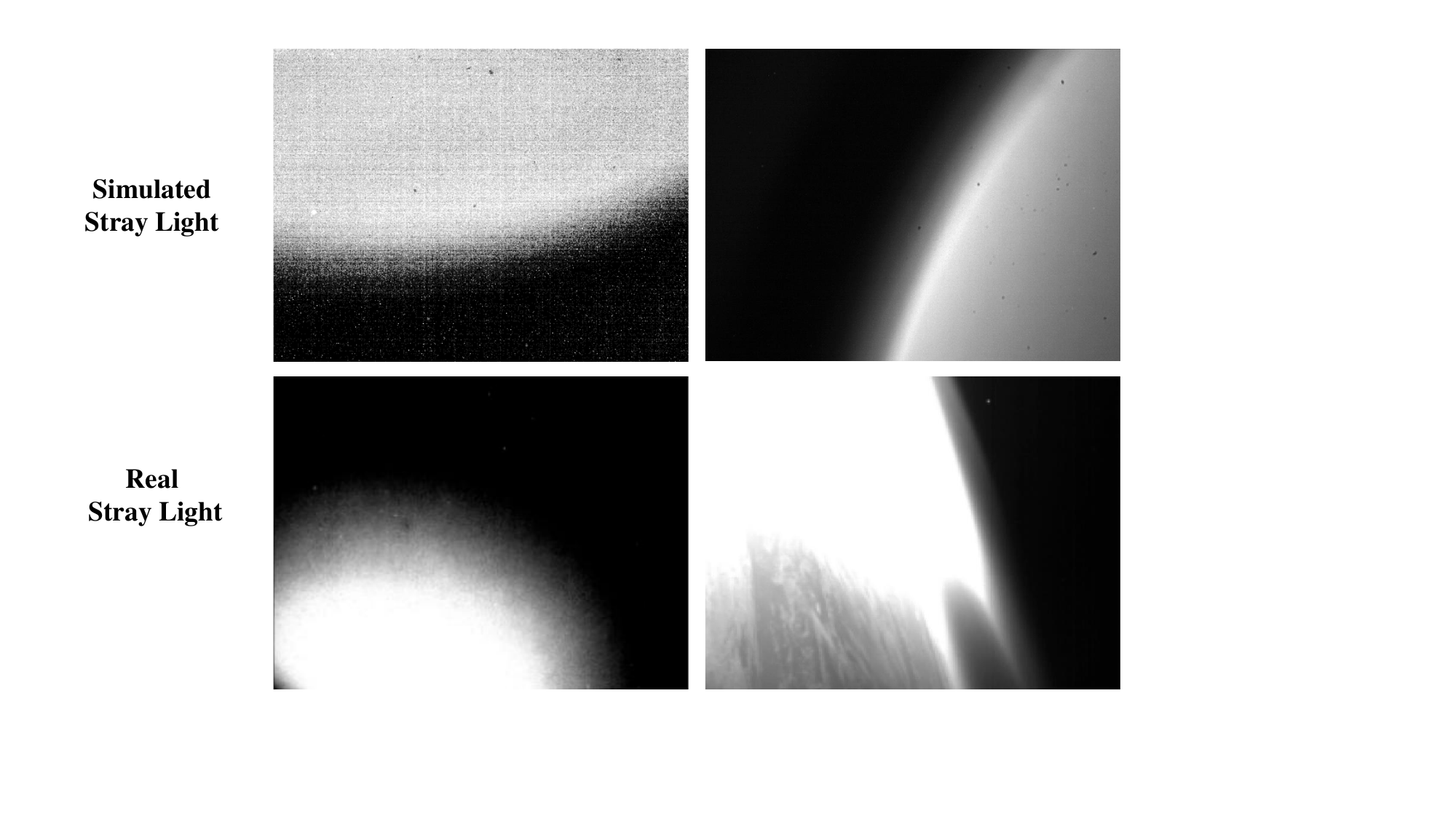}} 
  \caption{ \textcolor{DGreen}{Our simple setup effectively simulates realistic stray light patterns when compared to real star tracker images with moonlight/sunlight (first column) and earth-atmosphere light (second column) interference\cite{RealStray}.} }
  \label{fig:stray_light_compare}
\end{figure}

\begin{figure}
  \centerline{\includegraphics[trim={2.8cm 6.5cm 13cm 2cm},clip,width=20pc]{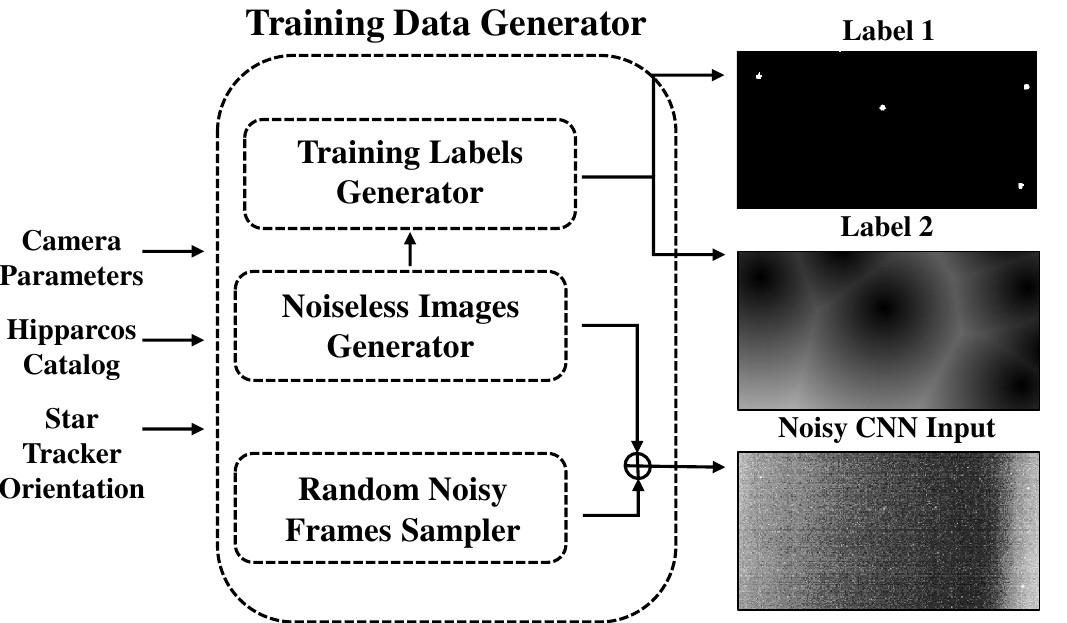}} 
  \caption{The noiseless simulated star images are merged with noisy frames containing stray light captured by actual cameras to create training inputs to CNN. 
  Additionally, two sets of training labels are generated: training labels 1, consisting of binary segmentation maps, and training labels 2, comprising distance maps.}
  \label{fig3:data}
\end{figure}

The pristine simulated star images are subsequently merged with noisy frames containing stray light captured by actual cameras, as illustrated in Fig. \ref{fig3:data}.
Current star detection studies typically apply Gaussian noise to simulated star images \cite{italy}\cite{SAA-UNet}.
Monakhova \textit{et al.} \cite{dance} demonstrated that simplistic Gaussian or Poisson-Gaussian noise models fail to adequately represent the frequently encountered non-Gaussian, non-linear, and sensor-specific noise inherent in low-light images.
Inspired by their observations, we acquire noisy dark frames from three distinct cameras by covering their lenses with lens caps.
For each camera, we captured 100 frames at exposure times of 100 ms, 300 ms, 500 ms, 700 ms, and 900 ms, respectively.
An additional set of 1000 frames is acquired by situating a camera within a darkroom environment and illuminating it with a flashlight from 12 distinct angles to simulate the presence of stray light.
\textcolor{DGreen}{As shown in Fig. \ref{fig:stray_light_compare}, our simple set up can simulate both moonlight/sunlight interference and earth-atmosphere light encountered by real satellite systems \cite{RealStray}.} 
During the training image generation process, a noisy frame is randomly selected from this pool of 2500 captured frames to be fused with a noiseless star image. 
Random flips are applied to the noisy frame to facilitate data augmentation.

As shown in Fig. \ref{fig3:data}, two distinct label types are produced for each training image: binary segmentation maps and distance maps. 
A binary segmentation map, mirroring the resolution of the training image, assigns each pixel a value of either ``one'' or ``zero.''
Pixels with a value of ``one'' denote star signals, while those with a value of ``zero'' represent background areas. 
Additionally, distance maps, sharing the same resolution as the training image, are segmented into Voronoi cells.
Each Voronoi cell $\mathcal{V}_k$ represents a cluster of pixels within the image $I$ that are nearest to a star $k$ with a centroid coordinate $C_{k}$. 
Mathematically, a Voronoi cell is defined as:
\begin{equation}
\mathcal{V}_k = \{ q \in I \: | \: \Vert q - C_{k} \Vert \leq \Vert q - C_{j} \Vert, \: \forall j \}
\end{equation}
Here, each pixel $q$ within a Voronoi cell $\mathcal{V}_k$ stores the distance $\Vert q - C_{k} \Vert$ from its center to the centroid of the corresponding star.
During training, a CNN is tasked with generating both segmentation and distance maps that accurately correspond to the labels, given a star image as input. 
This comprehensive training regimen enables the CNN to effectively learn the tasks of star detection and centroiding simultaneously.

\subsection{Star Detection and Centroiding Procedure}
    
    \begin{algorithm} 
        \caption{}
        \label{procedure}
        \begin{algorithmic}[1]
        \Procedure{Detect and Centoid}{$I_{in}$}      
            \State $S,D = f(I_{in})$ 
            \State $I_{out} = S \otimes  D$
            \While{$q \in I_{out}$ with $d<d_{th}$ exists}
                \State Find a coarse centroid pixel with $d<d_{th}$
                \State Put a $5 \times 5$ window around the pixel
                \For{$q_i$ within the window}
                    \State Obtain $(u_i, v_i)$ and $d_i$
                    \State remove $q_i$ from $I_{out}$ 
                \EndFor
                \State Obtain $A$ and $B$ using Eq. \ref{AX=B}
                \State $(u_c, v_c)$ = DGELSY($A,B$)
                \State store $(u_c, v_c)$ into a list
            \EndWhile 
          \State \Return the list of centroids
      \EndProcedure
  
    \end{algorithmic}
\end{algorithm}
Our star detection and centroiding procedure from image input to centroid output is described in algorithm \ref{procedure}. 
Given a captured star image $I_{in}$ as the input, neural network $f(I_{in})$ outputs both a segmentation map $S$ and a distance map $D$. 
To filter out the background, we perform an element-wise multiplication for both maps $I_{out} = S \otimes  D$.
To first find coarse centroid positions, we find every pixel $q \in I_{out}$ that has a distance $d$ value smaller than a threshold $d_{th}$.
A pixel $q$ with smaller $d$ indicates it is closer to the corresponding star centroid. 
In our experiments, we found that our method is not sensitive to the selection of $d_{th}$ and $d_{th}=0.5\sqrt{2}$ gives good results.
A $5 \times 5$ window is put around a coarse centroid pixel to obtain the coordinate $(u_i, v_i)$ and the distance value $d_i$ of every pixel $q_i$ within.

To find the coordinate $(u_c, v_c)$ of the corresponding star centroid, we first obtain the following equation
\begin{equation}
\label{Eq:distance}
    \begin{bmatrix}
        (u_1 - u_c)^2 + (v_1 -v_c)^2 \\
        (u_2 - u_c)^2 + (v_2 -v_c)^2 \\
        \vdots \\
        (u_n - u_c)^2 + (v_n -v_c)^2 \\
    \end{bmatrix}
    =
    \begin{bmatrix}
        d_1^2\\
        d_2^2\\
        \vdots\\
        d_n^2
    \end{bmatrix}
\end{equation}
We want to solve for the centroid coordinate using distances from a list of known coordinates, which is essentially a trilateration problem. Eq. \ref{Eq:distance} can be rewritten in the form of $Ax=B$
\begin{align}
    \label{AX=B}
    & \begin{bmatrix}
        2(u_n - u_1) & 2(v_n-v_1)\\
        2(u_n - u_2) & 2(v_n-v_2) \\
        \vdots\\
        2(u_n - u_{n-1}) & 2(v_n-v_{n-1})
    \end{bmatrix}   
    \begin{bmatrix}
        u_c \\
        v_c 
    \end{bmatrix} 
    = \nonumber \\
    & \begin{bmatrix}
        d_1^2 - d_n^2 - u_1^2 - v_1^2 + u_n^2 + v_n^2 \\
        d_2^2 - d_n^2 - u_2^2 - v_2^2 + u_n^2 + v_n^2 \\
        \vdots \\
        d_{n-1}^2 - d_n^2 - u_{n-1}^2 - v_{n-1}^2 + u_n^2 + v_n^2 
    \end{bmatrix}
\end{align}
We can obtain the centroid coordinates $(u_c, v_c)$ by passing matrices $A$ and $B$ into an off-the-shelf least-squares solver.
We use the DGELSY solver based on complete orthogonal factorization from the LAPACK library for its fast computation speed and numerical stability \cite{DGELSY}. 
After a region of pixels has been processed and its star centroid has been obtained, we remove all of these pixels from $I_{out}$ to avoid repetitive computations.
The process described above repeats until no pixel $q \in I_{out}$ with $d < d_{th}$ exists.

\subsection{CNN Implementation Details}
We implemented and compared four UNet-based CNN models, as detailed in Table \ref{CNNTable}: the vanilla UNet \cite{UNet}, MobileUNet \cite{MobileUNet}, ELUNet \cite{ELUNet}, and SqueezeUNet \cite{SqueezeUNet}. Each CNN model is designed with an input dimension of one for monochrome star images and an output dimension of two for segmentation and distance maps. The term ''Encoder Dim'' refers to the output channel dimension of the encoder blocks, with the dimensions of the decoder blocks adjusted accordingly. For instance, in the vanilla U-Net model with encoder dimensions of 64-128-256-512-1024, the decoder blocks have output dimensions of 512-256-128-2.
The original ELUNet employs bilinear interpolation for upsampling. However, our chosen edge AI processor does not support high-resolution bilinear upsampling operations. Consequently, we substituted all bilinear upsampling layers with transposed convolution layers in our implementations.
All CNN models were developed using PyTorch, and we based our implementations on existing public repositories of these models.

\begin{table}
\caption{CNN Models}
\label{CNNTable}
\tablefont
\centering
\begin{tabular}{|c|c|c|c|}
\hline
\textbf{Model} & \textbf{Input Dim} & \textbf{Encoder Dim} & \textbf{Out Dim} \\
\hline
UNet & 1 & 64-128-256-512-1024 & 2 \\ \hline
MobileUNet & 1 & 16-24-32-96-1280 & 2 \\ \hline
ELUNet & 1 & 8-16-32-64-128 & 2 \\ \hline
SqueezeUNet & 1 & 64-128-256-512-1024 & 2 \\
\hline
\end{tabular}
\end{table}
We employed binary cross-entropy loss, denoted as $L_S$, for training the segmentation map and mean squared error loss, denoted as $L_D$, for the distance map.
The total loss function is formulated as a weighted sum of these two individual loss terms
\begin{align}
    &L_S = \nonumber  \\
    & - mean \left( \sum_i^N S_i \otimes \log \hat S_i + \left( 1-S_i \right) \otimes \log \left( 1-\hat S_i \right) \right)\\
    &L_D = mean \left( \sum_i^N \left( \hat S_i \otimes \hat D_i - S_i \otimes D_i \right)^2 \right) \\
    &\textcolor{DGreen}{L_{total} = L_S + \alpha L_D}
\end{align}
Here, the $mean()$ function computes the average value of an array, $N$ represents the batch size, $S_i$ is the actual binary segmentation map, $\hat S_i$ is the estimated segmentation map with values ranging between zero and one, $D_i$ is the actual distance map, and $\hat D_i$ is the estimated distance map. Notation $\otimes$ denotes element-wise multiplication.
For our training procedure, we set \textcolor{DGreen}{$\alpha = 2.5$ }to balance the magnitude of the two loss terms. 

Our dataset consists of 2,500 images for training and an additional 500 images for evaluation.
CNN models undergo training for 100 epochs with a batch size set to 10 to leverage batch normalization layers effectively.
We employ the Adam optimizer with a weight decay of $5 \times 10^{-4}$ and an initial learning rate of $1\times 10^{-3}$.
The learning rate is decreased by a factor of two every 20 epochs to facilitate better convergence.
All training processes are conducted on Google Colab using a single NVIDIA V100 GPU.

\section{Experiments}
\subsection{Synthetic Data Test}
We begin our experimental evaluation by benchmarking our method against established algorithms using 500 synthetic test images.
To further assess the robustness of our approach, an additional set of 500 synthetic test images, contaminated by stray light, was employed.
Our evaluation encompasses various CNN models, including the vanilla UNet \cite{UNet}, SqueezeUNet \cite{SqueezeUNet}, ELUNet \cite{ELUNet}, and MobileUNet \cite{MobileUNet}.
For comparative analysis, we considered several existing star detection methods: Liebe's global threshold method \cite{Liebe}, the Weighted Iterative Threshold Method (WITM) \cite{WITM}, the ST-16 star tracker detection routine \cite{ST16}, and Sun \emph{et al.}'s local threshold method \cite{erosion}.
We also compared our method's centroiding performance to that of the center of gravity \cite{Liebe} and Gaussian grid algorithms \cite{Gaussian}.
The details of the CNN models and traditional methods are discussed in Section \rom{2}.

We employed precision, recall, and the F1 score as metrics to assess the detection performance. 
Precision measures the algorithm's capability to exclude false detections, while recall quantifies its ability to identify all stars. 
The F1 score offers a balanced evaluation by combining both precision and recall. 
\textcolor{DGreen}{Maintaining high recall is vital to ensure a sufficient number of stars are available for reliable identification, and more detected stars lead to better attitude determination accuracy in general. While modern star identification methods are robust to a reasonable number of false positives, excessive false detections can increase computational burden and potentially hinder real-time performance. Therefore, maintaining reasonable precision is still relevant.}

The mathematical expressions for these metrics are defined as
\begin{align} \label{detect_metrics}
    &Precision = \frac{TP}{TP+FP} \times 100 \\
    &Recall = \frac{TP}{TP+FN} \times 100 \\
    &F1 = 2 \times \frac{Precision \times Recall}{Precision + Recall}
\end{align}
In these formulas, ``TP'' denotes true positives, representing stars correctly identified by the star tracker. 
``FP'' indicates false positives, and ``FN'' signifies false negatives, which are stars not detected by the star tracker.
To evaluate centroiding accuracy, we employ the root-mean-square error (RMSE) between the actual centroid coordinates and the estimated coordinates.
\textcolor{DGreen}{Accurate centroid estimation is essential because it directly impacts both star identification and the calculation of precise orientation.}
Additionally, we document the computational time, the total number of multiply-accumulate operations (MACs) in giga, and the number of model parameters in millions for each CNN model. All CNN models are executed on an NVIDIA RTX2060M GPU. It is worth noting that the proposed least squares centroiding process runs efficiently, requiring only 4.49 milliseconds on an AMD Ryzen 7 4800H CPU.

Evaluation results are presented in Tables \ref{tab:synthetic_detect}, \ref{tab:synthetic_centroid}, \ref{tab:synthetic_straylight}, and \ref{tab:synthetic_speed}. 
Across both detection and centroiding assessments, all CNN models consistently outperformed traditional methods. 
We also found that the introduction of stray light had minimal impact on the CNN models' performance. 
We exclusively showcase the stray light performance of our proposed method, as existing detection methods struggle with these contaminated images.
Remarkably, despite having a considerably smaller model size and fewer computations, ELUNet achieves performance comparable to UNet. 
While MobileUNet exhibits slightly lower performance than other CNN models in both detection and centroiding, it distinguishes itself as the fastest method among the tested models.
On the other hand, despite its computational intensity, SqueezeUNet does not surpass ELUNet in performance. 
Consequently, due to their computational efficiency coupled with high performance, we exclusively focused on MobileUNet and ELUNet for the subsequent tests.

\begin{table}[hp]
    \caption{Detection Evaluation with Synthetic Images}
    \centering
    \begin{tabular}{|c|c|c|c|} \hline
        \textbf{Method} & \textbf{Precision} & \textbf{Recall} & \textbf{F1}\\ \hline
        UNet & 99.6 & 97.5 & 98.6 \\ \hline
        SqueezeUNet & 99.6 & 97.5 & 98.5 \\ \hline
            ELUNet & 99.7 & 97.5 & 98.6\\ \hline
        MobileUnet & 99.3 & 96.0 & 97.6\\ \hline
        Liebe's & 91.8 & 89.6 & 90.7\\ \hline
        WITM & 91.1 & 62.2 & 73.9\\ \hline
        ST16 & 93.4 & 83.6 & 88.2\\ \hline
        Sun \emph{et al.}'s & 81.1 & 93.9 & 87.0\\ \hline
    \end{tabular}
    \label{tab:synthetic_detect}
\end{table}

\begin{table}[hp]
    \caption{Centroiding Evaluation with Synthetic Images}
    \centering
    \begin{tabular}{|c|c|} \hline
        \textbf{Method} & \textbf{Centroid RMSE (pixels)} \\ \hline
        UNet & 0.1298 \\ \hline
        SqueezeUNet & 0.1382 \\ \hline
        ELUNet & 0.1363 \\ \hline
        MobileUnet & 0.1695 \\ \hline
        Center of Gravity & 0.6966 \\ \hline
        Gaussian Grid & 0.1922 \\ \hline
    \end{tabular}
    \label{tab:synthetic_centroid}
\end{table}

\begin{table}[hp]
    \caption{Evaluation with Straylight Synthetic Images}
    \centering
    \begin{tabular}{|c|c|c|} \hline
        \textbf{Method} & \textbf{F1} & \textbf{Centroid RMSE (pixels)} \\ \hline
        UNet & 97.3 & 0.1312  \\ \hline
        SqueezeUNet & 96.7 & 0.1394  \\ \hline
        ELUNet & 97.3 & 0.1348 \\ \hline
        MobileUnet & 96.6 & 0.1567 \\ \hline
    \end{tabular}
    \label{tab:synthetic_straylight}
\end{table}

\begin{table}[h]
    \caption{CNN Model Size and Computation Time}
    \centering
    \begin{tabular}{|c|c|c|c|} \hline
        \textbf{Model} & \textbf{Time (s)} & \textbf{MACs (G)} & \textbf{Params (M)} \\ \hline
        UNet & 0.113 & 256.25 & 31.04 \\ \hline
        SqueezeUNet & 0.067 & 70.06 & 2.64 \\ \hline
        ELUNet & 0.028 & 22.50 & 0.8 \\ \hline
        MobileUnet & 0.020 & 7.71 & 4.4 \\ \hline
    \end{tabular}
    \label{tab:synthetic_speed}
\end{table}

\subsection{Hardware Inference Time Test}
We continued evaluating our method using an Edge AI processor. Generally, space applications require hardware with low power consumption to accommodate a constrained power budget. The attitude update rate of a star tracker, in particular, necessitates hardware to possess at high processing speed to facilitate rapid inferences by the AI model. A few viable options on the market are identified by G. Furano \emph{et al.} \cite{edge}: Intel Myriad X, Google Coral Edge TPU, and Nvidia Jetson Nano. Ultimately, we selected the Google Coral Edge TPU for its low power consumption of 2 Watts, relatively high processing speed of 4 TOPS, and its readily available documentation and well-supported community.


To execute MobileUNet and ELUNet on the Edge TPU and measure the inference time, the model must first undergo quantization. This process transforms the high-bit representations of weights and activations in the original model into low-bit representations, thereby reducing the model’s inference time and enabling efficient execution on edge AI hardware. TinyNN \cite{quant} is employed to perform quantization on the model. Then, the model is compiled using the EdgeTPU compiler \cite{compiler}, resulting in an Edge TPU model. However, because the Edge TPU has limitations in terms of the model operations it supports \cite{compiler}, we substitute the resize bilinear operation in our model with the supported transpose convolution operation prior to compilation. The model is then ready to be loaded onto the Edge TPU for inference time measurement.


The Edge TPU is interfaced with a Raspberry Pi 4B over a USB 3.0 connection, and a sample test image file is loaded onto the device. Given that the model must initially be loaded into the Edge TPU’s memory, the first inference operation requires a longer duration to complete. To ensure the consistency of data collection, a set of twenty inferences is executed prior to any inference time measurements. Subsequently, a series of 100 inferences is performed on the sample test image and the time for each inference is measured. The mean and standard deviation of all the inference times are then computed. This procedure is replicated for both ELUNet and MobileUNet, and the results are consolidated in Table \ref{inference time}.
\begin{table}[H]
    \centering
    \caption{Edge TPU Inference Results}
    \begin{tabular}{|c|c|c|c|}
       \hline
       \textbf{Model}  & \textbf{Mean (ms)} & \textbf{$\sigma$ (ms)} & \textbf{Int8 File Size (MB)} \\
       \hline
       ELUNet  & 808.3532 & 83.3374 & 15.5 \\
       \hline
       MobileUNet & 265.5293 & 21.2383 & 6.7 \\
       \hline
    \end{tabular}
    \label{inference time}
\end{table}
MobileUNet performs exceptionally with a frequency of 3.77 Hz. ELUNet's mean inference time is under 1000 milliseconds with a frequency of 1.237 Hz. However, the performance of ELUNet is noticeably worse. This lower performance in terms of inference time can be ascribed to ELUNet having to perform an increased number of computations on higher-resolution feature maps, a consequence of its deep skip connections, despite it having fewer model parameters. 

\textcolor{DGreen}{
It is important to note that this was a preliminary investigation using simple static post-training quantization technique. While it effectively demonstrated the feasibility of real-time operation on an edge AI device, the resulting detection and centroiding performance were not desirable. To fully realize the potential of our algorithm on edge AI platforms, more advanced quantization techniques, such as quantization-aware training, will be necessary. These advanced techniques are beyond the scope of this current work, but they represent an important direction for future research and development.
}

\subsection{Static Hardware-in-the-loop Test}
\begin{figure}[hbt!]
    \centering
    \includegraphics[width=0.9\linewidth]{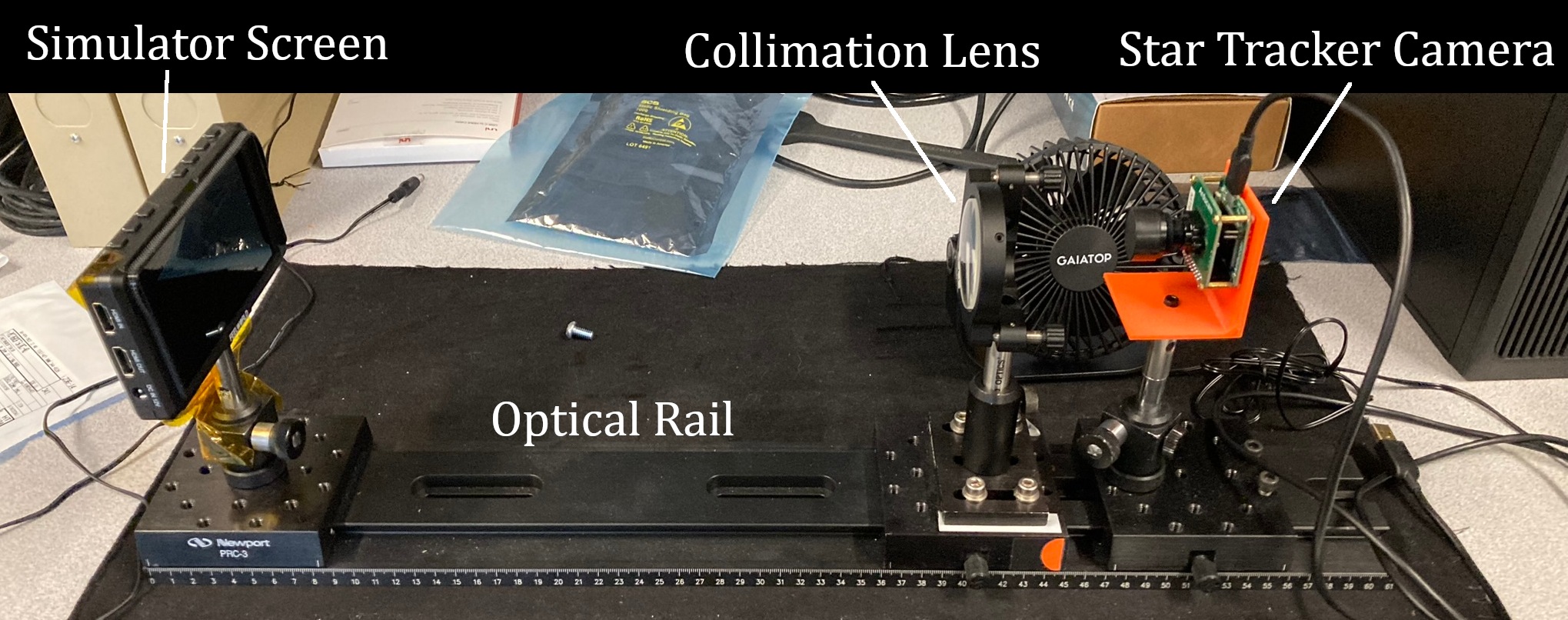}
    \caption{Starfield simulator allowed us to evaluate our method with real camera images. \textcolor{DGreen}{Components of the simulator are labeled accordingly.}}
    \label{fig:simulator}
\end{figure}

To gather data on the performance of our models, we constructed a starfield simulator. Creating our own starfield simulator enables us to easily control the conditions of the test and conduct them at any time. Specifically, our simulator system consists of a screen, a collimating lens, and a testing camera. The screen is a 5" diagonal 1920x1080 resolution IPS monitor and the achromatic collimation lens has a focal length of 350mm, a distance that lets the monitor fit nicely in the field of view of our camera. These components are mounted on optical posts, which slide along an optical rail to align and adjust their positions. The camera specifications are detailed in Table \ref{tab:STS-cam}. An image of the simulator setup is shown in Fig. \ref{fig:simulator}.

\begin{table}[hbt!]
    \caption{Camera Specifications}
    \centering
    \begin{tabular}{|c|c|} \hline
        \textbf{Sensor} & CMOS Monochrome Global Shutter\\ \hline
        \textbf{Resolution} & 640x480\\ \hline
        \textbf{Focal Length} & 16mm\\ \hline
        \textbf{Full Field of View} & 13.8x10.3 Degrees\\ \hline
        \textbf{F/ratio} & 1.2\\ \hline
    \end{tabular}
    \label{tab:STS-cam}
\end{table}
The simulator functions by displaying a set of stars on the screen from the Hipparcos Catalog limited to stars brighter than visual magnitude 6. The camera views this visual field through the collimating lens, which is placed to focus the starfield to infinity. The images captured by the camera are then fed into the star tracking algorithms for our tests. A cover lined with light-absorbing fabric is placed over the setup to ensure minimal light leakage into the simulator.
With this setup, there are some limitations. Since our screen has a limited spatial and luminance resolution, the positions and magnitude of stars are displayed with limited accuracy. The star vectors, on average, will have 12.9 arcseconds of angular distance error. The small dot pitch of our chosen monitor helps minimize this error. Based on lux measurements of the screen, we also estimate that the displayed stars have an apparent magnitude of 2.35.

With the simulator setup complete, we were able to run tests to compare our neural network and conventional baseline models. The first series involved a static starfield to measure performance, with our detection metrics being the precision, recall, and F1 scores defined in (\ref{detect_metrics}). To maintain consistency between tests, the simulator is calibrated by aligning the camera image center with the center of the monitor, establishing a fixed optical path. In addition, a fan is placed near the camera to provide cooling \textcolor{DGreen}{to the sensor} when desired, controlling the amount of thermal noise present in the camera images. \textcolor{DGreen}{The cooling from the fan reduced the sensor temperature from 29$^{\circ}$ Celsius to 24$^{\circ}$ Celsius in our tests.} Ten celestial fields of view were selected to provide test images for the algorithms. The camera integration time was fixed at 150 ms for all tests, and distortion correction was implemented using calculated OpenCV distortion coefficients. The parameters for each traditional method were kept the same for all tests; these were set to values with the best detection results for a real night sky image with 150 ms integration time. The results for the static detection are shown in Table \ref{tab:static-detection}.
\begin{table}
    \caption{Static Detection Results}
    \centering
    \begin{tabular}{|c|c|c|c|c|c|c|} 
        \cline{2-7} 
        \multicolumn{1}{c|}{}& \multicolumn{3}{c|}{\textbf{Cold}} & 
        \multicolumn{3}{c|}{\textbf{Hot}} \\ \hline
        \textbf{Model} & \textbf{Precision} & \textbf{Recall} & \textbf{F1} & \textbf{Precision} & \textbf{Recall} & \textbf{F1} \\ \hline
        ELUnet & 100 & 99.4 & 99.7 & 100 & 99.4 & 99.7  \\ \hline
        MobileUnet & 100 & 96.8 & 98.4 & 99.3 & 97.4 & 98.4 \\ \hline
        Liebe's & 94.9 & 47.5 & 63.3 & 100 & 43.2 & 60.4 \\ \hline
        WITM & 100 & 61.4 & 76.1 & 89.2 & 84.1 & 86.6 \\ \hline
        ST16 & 100 & 37.3 & 54.4 & 100 & 24.1 & 38.8 \\ \hline
        Sun \emph{et al.}'s & 100 & 43.0 & 60.2 & 98.9 & 59.2 & 74.1 \\ \hline
    \end{tabular}
    \label{tab:static-detection}
\end{table}

For both hot (noisy) and cold camera operating conditions, the neural networks performed comparably with near-perfect detection scores. Between the baseline methods, there is a notable difference between hot and cold operating conditions, with WITM and Sun \emph{et al.}'s methods improving in F1 score while the Liebe and ST16 methods had lower scores. We note that the traditional method performances are significantly worse here than when we tested them using synthetic data. We hypothesize that the parameters used for each method may not be fully optimized for these circumstances. However, this illustrates the robustness of our neural network models and how they do not need continuous recalibration to achieve good detection performance,  surpassing all tested traditional methods in precision, recall, and F1 scores.

Regarding the centroiding tests, we compared our neural networks to both the Center of Gravity and Gaussian Grid methods each paired with the WITM detection method, since it displayed the highest detection scores among the baseline methods. Since our static tests showed that our neural networks are not impacted by thermal noise and that the WITM model exhibited the highest detection performance in a hot camera operating condition, we tested centroiding accuracy with no cooling on the camera. To measure centroiding accuracy, we compared the difference in the angular separations of stars computed by each algorithm versus their true values displayed on the simulator screen. An average of 300 star pairs were compared for all algorithms. The static centroiding results are shown in Table \ref{tab:stat-cen}.

\begin{table}
    \caption{Static Centroiding Results}
    \centering
    \begin{tabular}{|c|c|c|c|c|} \hline
        \textbf{Model} & \textbf{Avg. Error (arcsec)} & \textbf{$\sigma$ (arcsec)}\\ \hline
        ELUnet & 56 & 50.8 \\ \hline
        MobileUnet & 57.9 & 53.6\\ \hline
        \makecell{WITM \\+ Gaussian Grid} & 58.2 & 49.2 \\ \hline
        \makecell{WITM \\+ Center of Gravity} & 71.2 & 61.1 \\ \hline
    \end{tabular}
    \label{tab:stat-cen}
\end{table}

Due to the limited-resolution error and any potential distortion calibration errors, the centroiding results may not represent the true capability of the algorithms. In addition, some stars near the far edge of the screen appear stretched due to optical distortion, which makes centroiding inaccurate for those stars. To counter this, we removed outlier stars that had average angular errors of greater than 200 arcseconds. Overall, since they were all tested in the same system and with the same calibration, the consistency enables comparison between the algorithms. Specifically, the neural networks are shown to have significantly improved centroiding accuracy to the Center of Mass method, while remaining competitive with the Gaussian Grid method. \textcolor{DGreen}{While the presence of close double stars has the potential to affect this accuracy, there exist only 17 pairs out of 5044 stars in the testing catalog exhibiting an angular separation of less than 0.6 arcminutes. As such, we expect minimal effect on our results due to the scarcity of these cases. Future improvements could include additional inclusions of double stars in the training data.}



Finally, we also tested performance when stray light is purposefully added to deteriorate the image quality. This was done by placing a flashlight inside the simulator, resulting in light flares in the captured images. The results for detection and centroiding accuracy are shown in Table \ref{tab:static-str-lgt}. Only our neural network models are tabulated for stray light results, as all other conventional models failed to identify any stars and were plagued with false positives when exposed to the same lighting conditions.

\begin{table}[hbt!]
    \caption{Stray Light Performance}
    \centering
    \begin{tabular}{|c|c|c|c|c|c|} \hline
        \textbf{Model} & \textbf{Precision} & \textbf{Recall} & \textbf{F1} & \textbf{Avg Error} & \textbf{$\sigma$}\\ \hline
        ELUnet & 94.4 & 96.8 & 95.6 & 60.5 & 51.9\\ \hline
        MobileUnet & 87.9 & 97.4 & 92.4 & 61.2 & 50.3 \\ \hline
    \end{tabular}
    \label{tab:static-str-lgt}
\end{table}

Even with stray light contamination, our proposed neural network methods display only slightly worse detection metrics while all other traditional models displayed a multitude of false positives. While the precision scores of the models are slightly less due to the appearance of some false positives, the neural network models are still able to consistently and correctly identify true stars in heavily contaminated images, as evidenced by their precision, recall, and F1 scores. Our models also retained their high accuracy under these conditions, averaging only about 4 arcseconds worse than the non-contaminated static tests.





\subsection{Dynamic Hardware-in-the-loop Test}

\begin{figure}[!hbt]
  \centerline{\includegraphics[width=19.5pc]{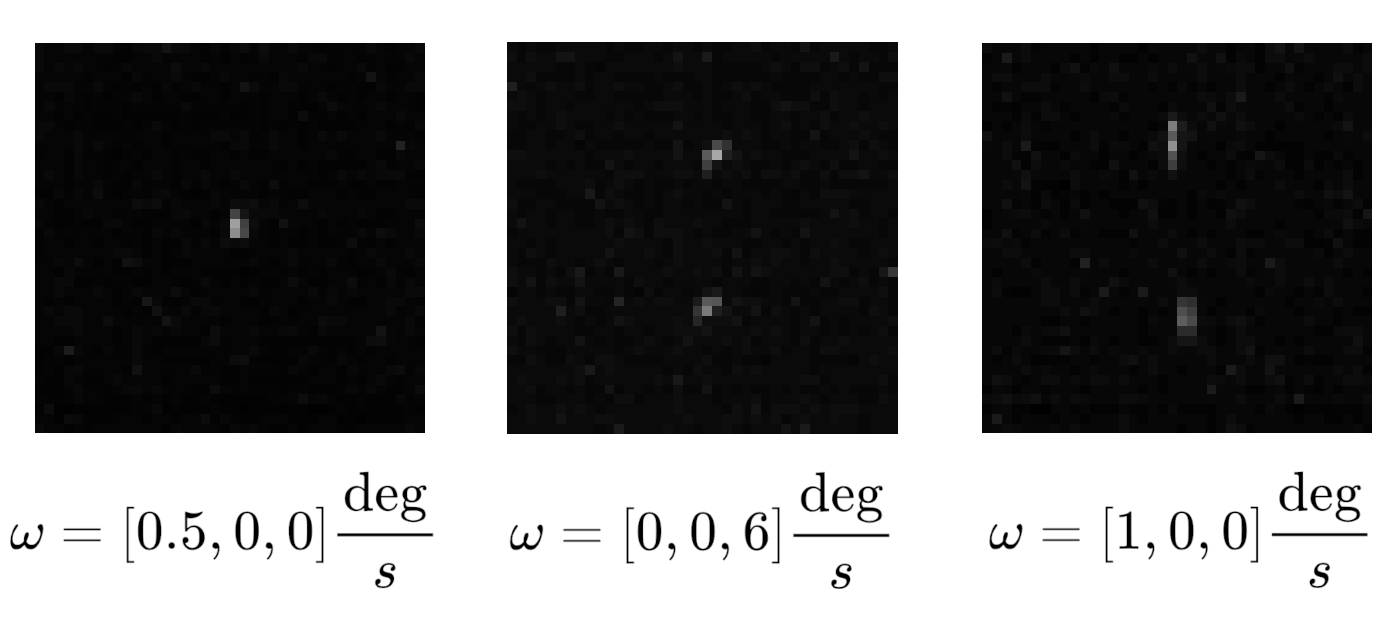}} 
    \caption{\textcolor{DGreen}{Star camera images showing stars during dynamic testing (enhanced for visibility). It is seen that higher rotation rates produce more elongated star pixels in the starfield images.}}
  \label{fig:rotationrates}
\end{figure}

The second series of testing involved a rotating starfield to evaluate detection performance. The simulator is calibrated in the same manner as the static tests. Since the results of the static tests demonstrated the quality of performance with varying levels of thermal noise, the dynamic tests were only conducted at one temperature. Ten fields of view (the same as the static tests) were chosen for the starfields. For each starfield, three separate rates of rotation were used to dictate the manner in which the displayed starfield moved across the monitor. Each method (ELUnet, MobileUnet, WITM, and Sun \emph{et al.}'s) was tested with a rotating starfield a total of thirty times. Fig. \ref{fig:rotationrates} demonstrates the star image quality depending on rotation rate. A random time was chosen to capture the stills used for analysis. By comparing the number of stars detected in the captured images to the simulated images and applying the metrics in Eq. \ref{detect_metrics}, we were able to come up with an F1 score that measures the accuracy of each method in detecting stars. 

The results for dynamic detection for $\omega$=[0.5, 0, 0], $\omega$=[0, 0, 6], and $\omega$=[1, 0, 0] are shown in Table \ref{tab:dyn-det}. For all three rotation rates, our proposed two neural network methods perform the best with F1 scores above 99. The WITM method achieved slightly lower scores due to having more false negatives. Nonetheless, it still performed significantly better than Sun \emph{et al.}'s method, which struggled notably in detecting faint stars.

\begin{table*}[!hbt]
\caption{Dynamic Detection Results}
\centering
\begin{tabular}{|c|ccc|ccc|ccc|}
\hline
 & \multicolumn{3}{c|}{\textbf{$\omega$={[}0.5, 0, 0{]}}} & \multicolumn{3}{c|}{\textbf{$\omega$={[}0, 0, 6{]}}} & \multicolumn{3}{c|}{\textbf{$\omega$={[}1, 0, 0{]}}} \\ \hline
\textbf{Model} & \multicolumn{1}{c|}{\textbf{Precision}} & \multicolumn{1}{c|}{\textbf{Recall}} & \textbf{F1} & \multicolumn{1}{c|}{\textbf{Precision}} & \multicolumn{1}{c|}{\textbf{Recall}} & \textbf{F1} & \multicolumn{1}{c|}{\textbf{Precision}} & \multicolumn{1}{c|}{\textbf{Recall}} & \textbf{F1} \\ \hline
ELUnet & \multicolumn{1}{c|}{100} & \multicolumn{1}{c|}{100} & 100 & \multicolumn{1}{c|}{100} & \multicolumn{1}{c|}{100} & 100 & \multicolumn{1}{c|}{99.3} & \multicolumn{1}{c|}{100} & 99.9 \\ \hline
MobileUnet & \multicolumn{1}{c|}{100} & \multicolumn{1}{c|}{99.3} & 99.65 & \multicolumn{1}{c|}{100} & \multicolumn{1}{c|}{100} & 100 & \multicolumn{1}{c|}{100} & \multicolumn{1}{c|}{100} & 100 \\ \hline
WITM & \multicolumn{1}{c|}{99.2} & \multicolumn{1}{c|}{80.1} & 88.6 & \multicolumn{1}{c|}{99.2} & \multicolumn{1}{c|}{84.6} & 91.3 & \multicolumn{1}{c|}{98.6} & \multicolumn{1}{c|}{91.3} & 94.8 \\ \hline
Sun \emph{et. al's} & \multicolumn{1}{c|}{98.4} & \multicolumn{1}{c|}{41.2} & 58.1 & \multicolumn{1}{c|}{100} & \multicolumn{1}{c|}{38} & 55.1 & \multicolumn{1}{c|}{100} & \multicolumn{1}{c|}{50.3} & 67.0 \\ \hline
\end{tabular}
\label{tab:dyn-det}
\end{table*}

\begin{table*}[!hbt]
\caption{Dynamic Centroiding Results}
\centering
\begin{tabular}{|c|cc|cc|cc|}
\hline
 & \multicolumn{2}{c|}{\textbf{$\omega$={[}0.5, 0, 0{]}}} & \multicolumn{2}{c|}{\textbf{$\omega$={[}0, 0, 6{]}}} & \multicolumn{2}{c|}{\textbf{$\omega$={[}1, 0, 0{]}}} \\ \hline
\textbf{Model} & \multicolumn{1}{c|}{\textbf{Error} \newline \textbf{(arcsec)}} & \textbf{$\sigma$ (arcsec)} & \multicolumn{1}{c|}{\textbf{Error (arcsec)}} & \textbf{$\sigma$ (arcsec)} & \multicolumn{1}{c|}{\textbf{Error (arcsec)}} & \textbf{$\sigma$(arcsec)} \\ \hline
ELUnet & \multicolumn{1}{c|}{69.59} & 61.02 & \multicolumn{1}{c|}{72.17} & 61.73 & \multicolumn{1}{c|}{59.3} & 46.6 \\ \hline
MobileUnet & \multicolumn{1}{c|}{62.56} & 52.52 & \multicolumn{1}{c|}{69.03} & 62.53 & \multicolumn{1}{c|}{69.1} & 52.5 \\ \hline
WITM + Gaussian Grid & \multicolumn{1}{c|}{69.82} & 53.69 & \multicolumn{1}{c|}{74.75} & 61.62 & \multicolumn{1}{c|}{82.72} & 56.37 \\ \hline
\end{tabular}
\label{tab:dyn-cen}
\end{table*}

The second phase of dynamic testing involved computing centroid accuracy by comparing the actual star angular distances from the star catalog with the angular distances computed by the algorithms. This evaluation was conducted across three different starfields, each featuring three distinct rates of rotation. This resulted in nine test images for each method (MobileUnet, ELUnet, and WITM Gaussian), each containing multiple stars. Similar to the static centroiding tests, stars with an average angular error of greater than 200 arcseconds were considered outliers and removed. The results for all three rates of rotation are presented in Table \ref{tab:dyn-cen}. In all cases, the neural networks had a lower error than the WITM Gaussian method. Despite not being trained on dynamic images, the proposed method demonstrates robustness for star images with motion blur, consistently producing accurate centroids.

\subsection{Night Sky Testing}
While hardware-in-the-loop tests offer control over temperature and rotation rates, our current hardware setup has limitations. It can only provide a coarse estimation of accuracy performance and lacks the capability to simulate faint stars. 
To achieve a more comprehensive evaluation, we performed night sky testing at Middle Fork River Forest Preserve, IL, U.S.
The camera, detailed in Table \ref{tab:STS-cam}, was mounted on a tripod and aimed near the zenith with its exposure time set to 500 ms. 
To align with standard practices \cite{Liebe}, the inertial-to-body attitude outputs were converted to 3-2-1 Euler angles representing right ascension, declination, and roll.
The right ascension would be expected to change linearly at Earth's sidereal rate, while the declination and roll remain constant.
The cross-boresight and around-boresight accuracy of a star tracker can be modeled as the standard deviations of the declination and roll. 
For star identification, we employed the geometric voting algorithm \cite{geometricVoting}, and the singular value decomposition method \cite{svd} was utilized for attitude determination. Recorded videos of the test results can be found at our project website \href{https://hongruizhao.github.io/CNNStarDetectCentroid/}{https://hongruizhao.github.io/CNNStarDetectCentroid/}.

\begin{figure}
  \centerline{\includegraphics[width=20pc]{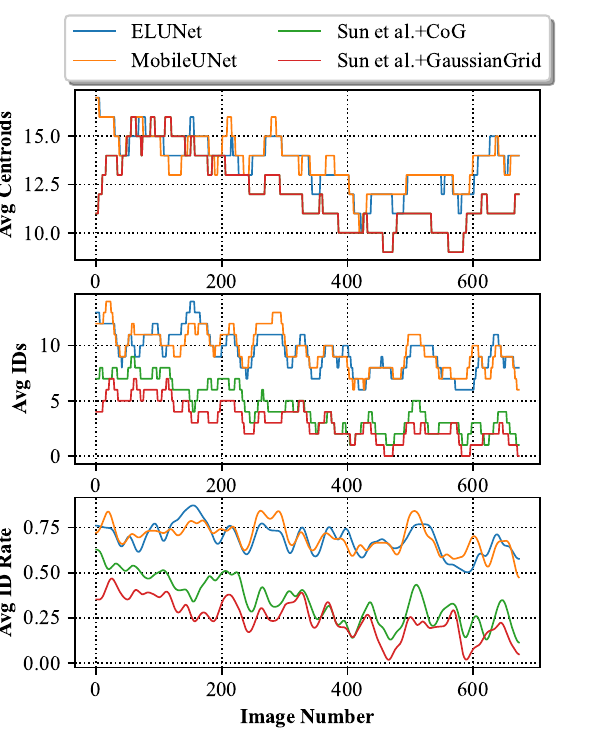}} 
  \caption{Detection and identification performance in the first night sky test. With a similar number of star centroids detected, the proposed method has higher star identification rates compared to the conventional methods.}
  \label{fig:nightSky}
\end{figure}

We recorded two 10-minute videos to evaluate the detection and accuracy of our method.  Results are presented in Table \ref{tab:nightSky}. Among conventional methods, Sun \emph{et al.}'s local threshold approach demonstrated the best detection performance and served as our comparison benchmark. Overall, both CNN models outperformed traditional approaches, with ELUNet showing a slight edge over MobileUNet. As illustrated in Fig. \ref{fig:nightSky}, while our method and the conventional baseline exhibit similar detection performance, our method achieved a significantly higher star identification rate due to its superior centroiding accuracy. 

\begin{table}
    \caption{Night Sky Test}
    \centering
    \begin{tabular}{|c|c|c|c|c|}  
        \cline{2-5} 
        \multicolumn{1}{c|}{}& \multicolumn{2}{c|}{\makecell{\textbf{Cross-Boresight} \\ \textbf{(arcsecond 1-$\sigma$)}}} & 
        \multicolumn{2}{c|}{ \makecell{\textbf{Around Boresight} \\ \textbf{(arcsecond 1-$\sigma$)}} } \\ \hline
        \textbf{Model} & \textbf{Test 1} & \textbf{Test 2} & \textbf{Test 1} & \textbf{Test 2} \\ \hline
        ELUnet & 12.92 & 13.79 & 76.95 & 151.90  \\ \hline
        MobileUnet & 12.93 & 15.44 & 81.49 & 146.88\\ \hline
        \makecell{Sun \emph{et al.}'s \\+ Center of Gravity} & 21.26 & 22.99 & 251.72 & 290.84 \\ \hline
        \makecell{Sun \emph{et al.}'s \\+ Gaussian Grid} & 19.84 & 19.34 & 221.09 & 253.34 \\ \hline
    \end{tabular}
    \label{tab:nightSky}
\end{table}

\begin{figure}
  \centerline{\includegraphics[trim={7.6cm 6.3cm 5.5cm 4.5cm},clip,width=20pc]{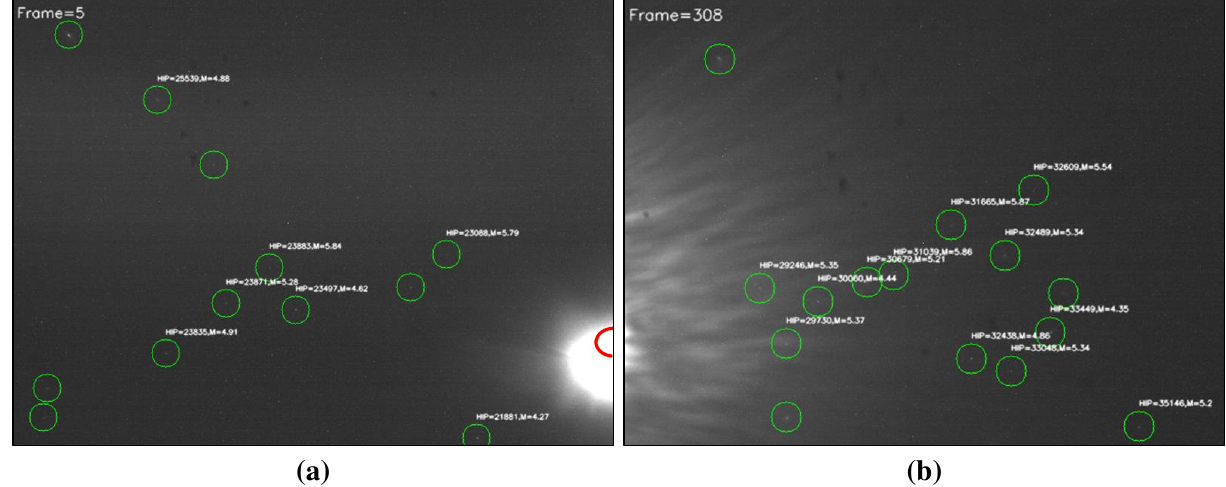}} 
  \caption{Despite (a) the presence of the moon within the field-of-view \textcolor{DGreen}{(a stray light pattern not present in our training data)} or (b) severe lens flare, our method leads to accurate star detection and identification. \textcolor{DGreen}{The only fake star detected is highlighted in red. This result highlights our method's capacity to handle diverse and unseen stray light conditions.}}
  \label{fig:moon}
\end{figure}

To assess the impact of stray light, we first recorded a 5-minute video under standard conditions, immediately followed by a subsequent 5-minute video with stray light introduced via a flashlight. 
The results are presented in Table \ref{tab:nightSkyStraylight}. 
While both models are affected by stray light, their accuracy remains comparable to typical commercial CubeSat star trackers. 
Under normal conditions, ELUNet and MobileUNet perform identically. 
However, when exposed to stray light, ELUNet demonstrated greater robustness. 
Fig. \ref{fig:moon} showcases visual examples of ELUNet prcoessing results of images affected by stray light and those with the moon in the field-of-view. 
Detected stars are highlighted with green circles, while identified stars display their information at the top. 
Even when confronted with significant lens flare or the moon's interference, our method reliably rejected false detections and precisely computed star centroids, leading to successful star identification.

\begin{table}
    \caption{Night Sky Test, Stray light}
    \centering
    \begin{tabular}{|c|c|c|c|c|}  
        \cline{2-5} 
        \multicolumn{1}{c|}{}& \multicolumn{2}{c|}{\makecell{\textbf{Cross-Boresight} \\ \textbf{(arcsecond 1-$\sigma$)}}} & 
        \multicolumn{2}{c|}{ \makecell{\textbf{Around Boresight} \\ \textbf{(arcsecond 1-$\sigma$)}} } \\ \hline
        \textbf{Model} & \textbf{\makecell{w/o\\Straylight}} & \textbf{Straylight} & \textbf{\makecell{w/o\\Straylight}} & \textbf{Straylight} \\ \hline
        ELUnet & 8.07 & 20.25 & 85.85 & 191.73  \\ \hline
        MobileUnet & 8.07 & 23.09 & 85.85 & 216.96\\ \hline

    \end{tabular}
    \label{tab:nightSkyStraylight}
\end{table}

\section{Conclusion}
In this article, we introduced a CNN-based method for real-time star detection and centroiding in star tracker processing. 
We generated training images by blending synthetic ``clean'' star images with authentic sensor noise and stray light. 
The CNN model yields a binary segmentation map for star detection and a distance map for centroiding. 
By incorporating distance data with pixel coordinates, centroid calculation was reformulated as a trilateration problem, which is solved using the least squares method. 
We conducted synthetic image tests, hardware-in-the-loop tests, and night sky tests to thoroughly assess our method against existing approaches. 
Our method not only surpassed existing methods in detection and centroiding performance but also exhibited remarkable robustness across varying thermal, rotation, and stray light conditions. 
\textcolor{DGreen}{We found that the MobileUNet model exhibited the best performance-to-computation-time trade-off, and is the most suitable for real time star tracking applications.}
The model operates with under 300 ms latency on a low-power Google Coral TPU. 

\textcolor{DGreen}{For future works, investigating the impact of aberrations and incorporating asymmetric patterns in the training data could further enhance performance.}
Another potential enhancement could involve training the CNN to process star trails resulting from higher rotation rates.
\textcolor{DGreen}{To bring our algorithm closer to real edge AI applications in space, we need to explore more advanced quantization techniques to minimize performance degradation.}
Moving forward, our research will progress to constructing a prototype star tracker to further validate our method in an actual flight-qualified setup.


\begin{IEEEbiography}
    [{\includegraphics[width=1in,height=1.25in,clip,keepaspectratio]{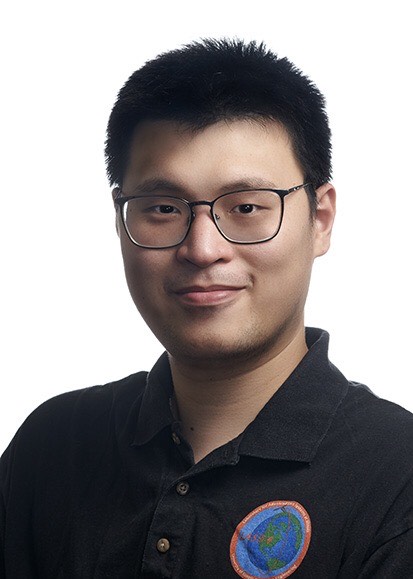}}]{Hongrui Zhao}
    was born in China in 1995.
    He holds an M.S in aerospace engineering from the University of Illinois Urbana Champaign (UIUC), 2020, and a B.S. in aerospace engineering from Nanjing University of Aeronautics and Astronautics, 2018.
    
    He is a PhD candidate in aerospace engineering at UIUC. 
    He is the GNC and star tracker research lead at the Laboratory for Advanced Space Systems at Illinois. 
    He has worked on multiple UIUC CubeSat missions as a GNC engineer, including CubeSail, SASSI2, and CAPSat. 
    He has worked as a GNC engineer intern at CU Aerospace from 2022 to 2023. 
    Currently, his research focuses on applying artificial intelligence techniques to address challenges in space applications. 
\end{IEEEbiography}

\begin{IEEEbiography}
    [{\includegraphics[width=1in,height=1.25in,clip,keepaspectratio]{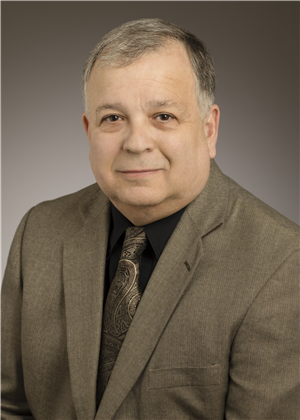}}]{Michael F. Lembeck} 
    was born in Peoria, IL. He has B.S. (1980), M.S.(1981), and Ph.D. (1991) degrees in Aeronautical and Astronautical Engineering from the University of Illinois at Urbana-Champaign (UIUC).
    
    He is a Clinical Associate Professor and Director of the Laboratory for Advanced Space Systems at Illinois (LASSI) providing students with opportunities to design, build, and fly experimental CubeSat satellites. He has led or worked on multiple government and commercial spaceflight programs, including JPL’s Galileo Jupiter Orbiter and Space Industries, Inc.’s Wake Shield Facility-03 flown on STS-80, Orbital Sciences’ OrbView-3 and OrbView-4/Warfighter satellites, and he was the Requirements Division Director for the Exploration Systems Mission Directorate at NASA Headquarters.
    
    Dr. Lembeck is an Associate Fellow of the American Institute of Aeronautics and Astronautics (AIAA). 

\end{IEEEbiography}

\begin{IEEEbiography}[{\includegraphics[width=1in,height=1.25in,clip,keepaspectratio]{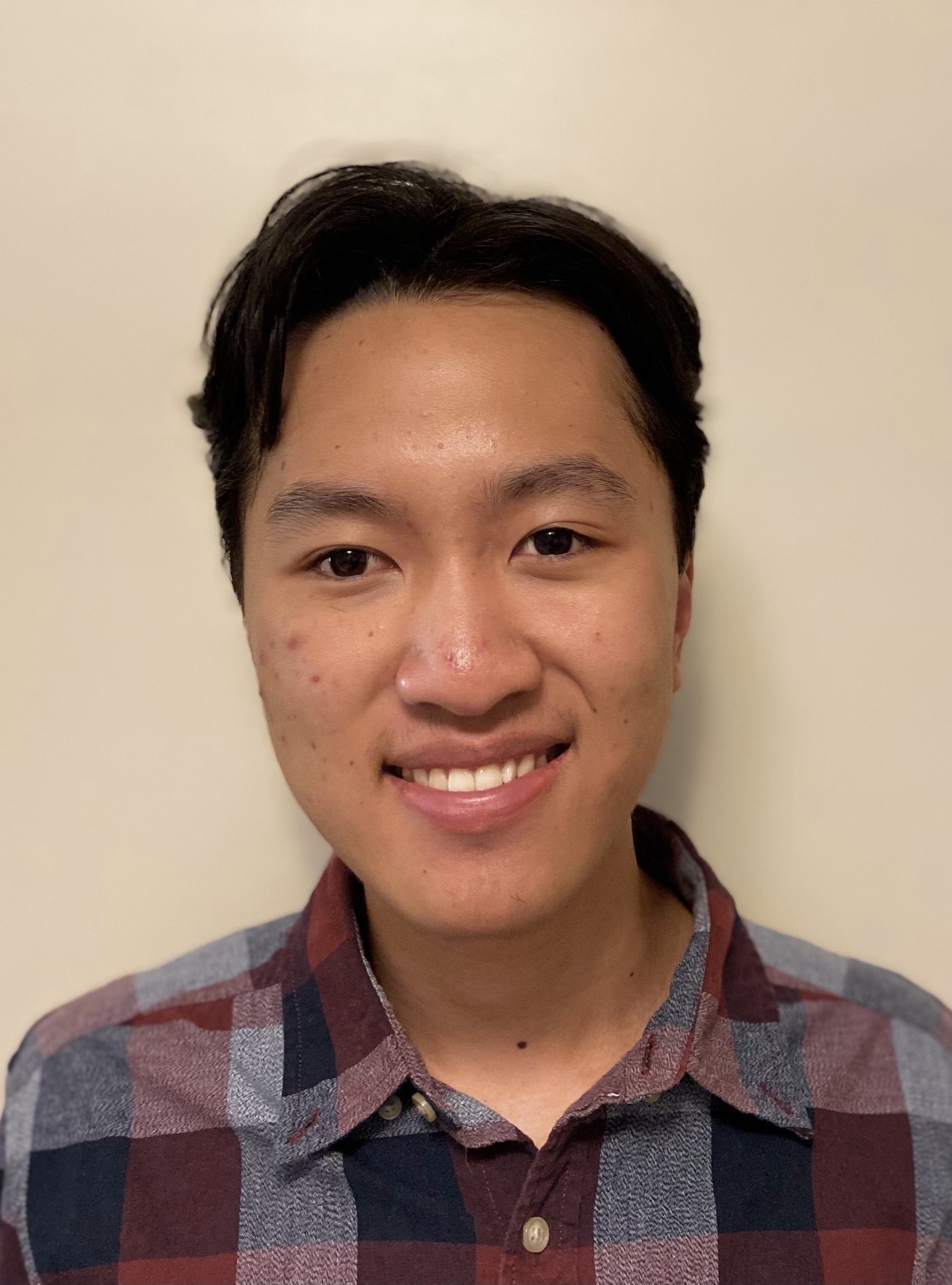}}]{Adrian Zhuang} is a third-year undergraduate student pursuing a B.S. degree in aerospace engineering at the University of Illinois at Urbana-Champaign.
He is performing research at the Laboratory for Advanced Space Systems at Illinois on star tracker systems and test systems.

\end{IEEEbiography}%

\begin{IEEEbiography}
    [{\includegraphics[width=1in,height=2in,clip,keepaspectratio]{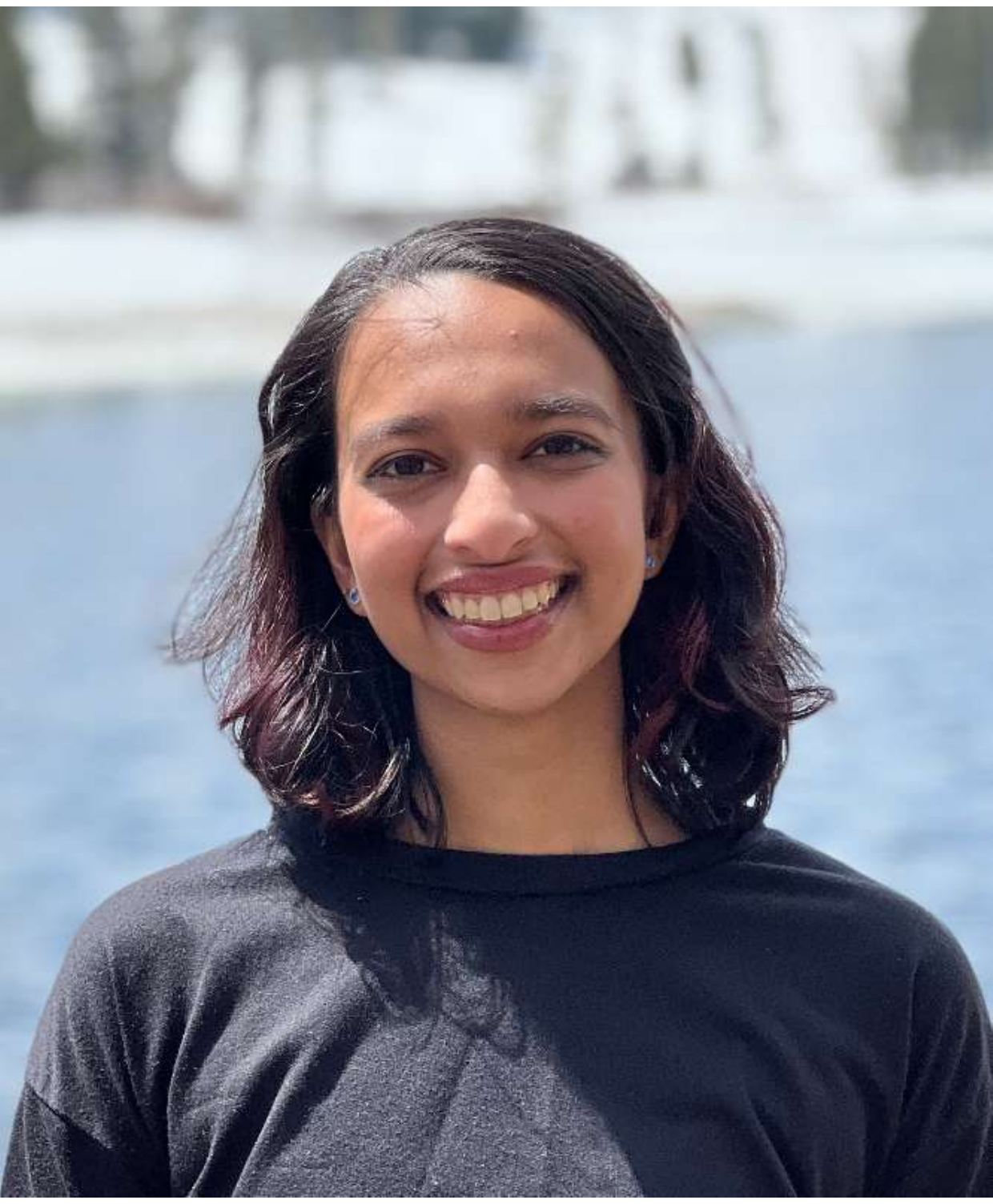}}]{Riya Shah} is a third-year undergraduate student studying aerospace engineering at the University of Illinois at Urbana-Champaign. She works at the Laboratory for Advanced Space Systems at Illinois (LASSI) on the star-tracker research project. 
\end{IEEEbiography}

\begin{IEEEbiography}[{\includegraphics[width=1in,height=1.25in,clip,keepaspectratio]{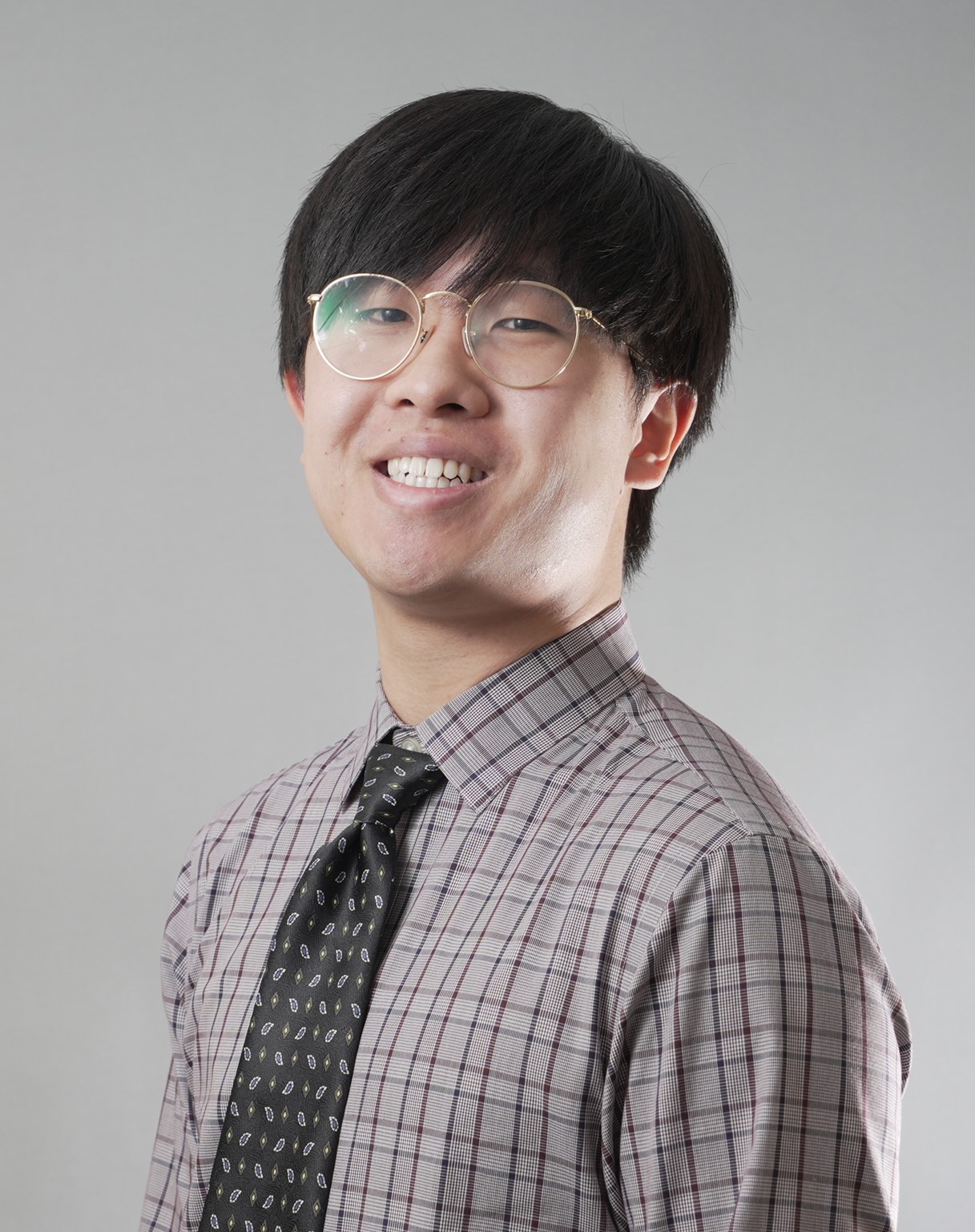}}]{Jesse Wei} is a third-year undergraduate student studying aerospace engineering at the University of Illinois Urbana-Champaign. He is working for the Laboratory for Advanced Space Systems at Illinois on the star-tracker research project.
\end{IEEEbiography}
\end{document}